%% file: main.tex
\ifdefined\pdftexversion\pdfoutput=1\fi
\documentclass[runningheads]{llncs}

\usepackage{eccv}

\usepackage{eccvabbrv}

\usepackage{graphicx}
\usepackage{booktabs}
\usepackage{multirow}

\usepackage[accsupp]{axessibility}%

\newcommand{\model}{GeoV2V}
\newcommand{\dataset}{Para4D}

\usepackage{hyperref}
\hypersetup{pdftitle={GeoV2V: Geometry-Grounded Video Diffusion Model for Driving Scene Generation},pdfauthor={Yuchen Xi, Yuhan Guo, Chenwei Hou, Zixu Liu, Jin Fang, Jason Liu, Ruigang Yang},pdfsubject={Author accepted manuscript, ECCV 2026, DOI 10.1007/978-3-032-37044-0\_36}}

\usepackage{orcidlink}

\begin{document}

\title{GeoV2V: Geometry-Grounded Video Diffusion Model for Driving Scene Generation}

\titlerunning{GeoV2V}

\author{
Yuchen Xi\inst{1}$^{\ast}$\orcidlink{0009-0008-0595-1151} \and
Yuhan Guo\inst{1}$^{\ast}$\orcidlink{0009-0008-5469-0269} \and
Chenwei Hou\inst{2}\orcidlink{0009-0000-8232-5199} \and
Zixu Liu\inst{3}\orcidlink{0009-0002-1821-7745} \and
Jin Fang\inst{1}\orcidlink{0000-0002-7947-6807} \and
Jason Liu\inst{4}$^{\ddagger}$\orcidlink{0009-0008-3689-986X} \and
Ruigang Yang\inst{1}$^{\dagger}$\orcidlink{0000-0001-5296-6307}
}

\authorrunning{Y.~Xi et al.}

\institute{
Global Institute of Future Technology, Shanghai Jiao Tong University, China
\and
Global College, Shanghai Jiao Tong University, China
\and
School of Mechanical Engineering, Shanghai Jiao Tong University, China
\and
Shanghai Pudong Development Bank, China
}

\maketitle

\begingroup
\renewcommand{\thefootnote}{}
\footnotetext{\textsuperscript{$\ast$} Equal contribution. Email: \email{yuchen.x@sjtu.edu.cn, tippyguo@sjtu.edu.cn}}
\footnotetext{\textsuperscript{$\dagger$} Corresponding author. Email: \email{ryang2@sjtu.edu.cn}}
\footnotetext{\textsuperscript{$\ddagger$} Work done while at the Global Institute of Future Technology, Shanghai Jiao Tong University.}
\footnotetext{\scriptsize This version of the contribution has been accepted for publication, after peer review, but is not the Version of Record and does not reflect post-acceptance improvements, or any corrections. The Version of Record is available online at: \url{https://doi.org/10.1007/978-3-032-37044-0_36}. Use of this Accepted Version is subject to the publisher's Accepted Manuscript terms of use: \url{https://www.springernature.com/gp/open-research/policies/accepted-manuscript-terms}.}
\endgroup

\begin{abstract}
Novel view synthesis under trajectory changes is essential for autonomous driving simulation, yet existing methods struggle to generate consistent videos when extrapolating to unseen viewpoints such as lane shifts. Reconstruction-based approaches maintain geometric consistency but degrade in visual quality under large viewpoint changes, while reconstruct-then-restore methods often produce visually plausible frames yet suffer from geometric and temporal inconsistencies. We present \textbf{GeoV2V}, a geometry-grounded video-to-video diffusion framework that synthesizes temporally coherent driving videos along novel trajectories by integrating LiDAR and depth-derived geometric priors with full-video conditioning in the Wan 2.1 backbone. A key challenge in learning cross-lane transformations is the lack of synchronized multi-trajectory supervision in real-world datasets. To address this, we also introduce \textbf{Para4D}, a synthetic dataset capturing synchronized multi-lane driving videos within dynamic scenes, which provides explicit supervision for lane-shift generation and serves as a strong training prior for geometry-consistent view synthesis. Experiments on Waymo, nuScenes, and Para4D demonstrate state-of-the-art performance in visual quality, geometric consistency, and robustness to complex lighting dynamics under large trajectory shifts. \noindent{Project page:} \url{https://xiyuche.github.io/GeoV2V/}

\end{abstract}

\input{sec/1_introduction}
\input{sec/2_related}
\input{sec/3_dataset}

\input{sec/4_method}
\input{sec/5_experiment}
\input{sec/6_conclusion}

\bibliographystyle{splncs04}
\bibliography{main}
\end{document}

%% file: sec/1_introduction.tex
\section{Introduction}

Generating driving videos from novel viewpoints is an important capability for autonomous driving simulation and world modeling. In particular, synthesizing views along unseen trajectories—such as lane changes—allows simulation systems to predict the visual consequences of alternative actions taken by the ego vehicle. Compared with standard view interpolation, lane-shift view synthesis requires extrapolating to viewpoints that deviate laterally from the recorded trajectory while preserving the spatial layout of the scene and the temporal dynamics of surrounding traffic. Achieving both geometric consistency and temporal coherence under such trajectory shifts remains a challenging problem.

\begin{figure}[t]
\centering
\begin{subfigure}{\linewidth}
    \centering
    \includegraphics[width=\linewidth]{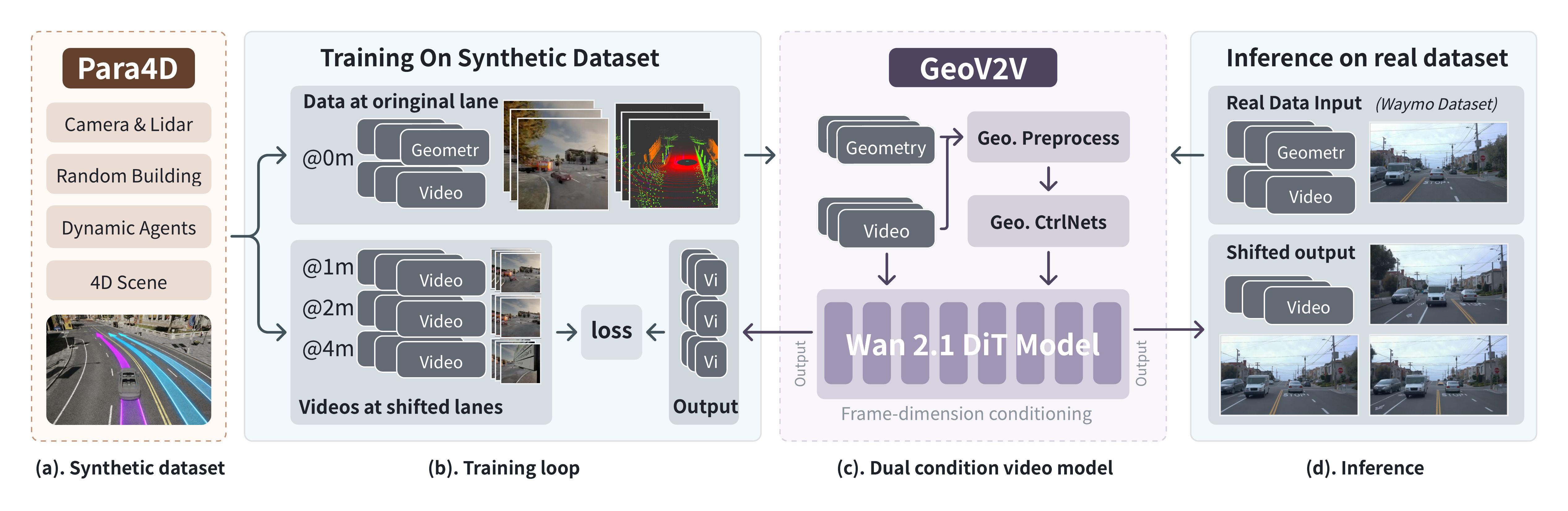}
    \caption{Overview of the GeoV2V pipeline for lane-shifting novel view synthesis.}
\end{subfigure}

\begin{subfigure}{\linewidth}
    \centering
    \includegraphics[width=\linewidth]{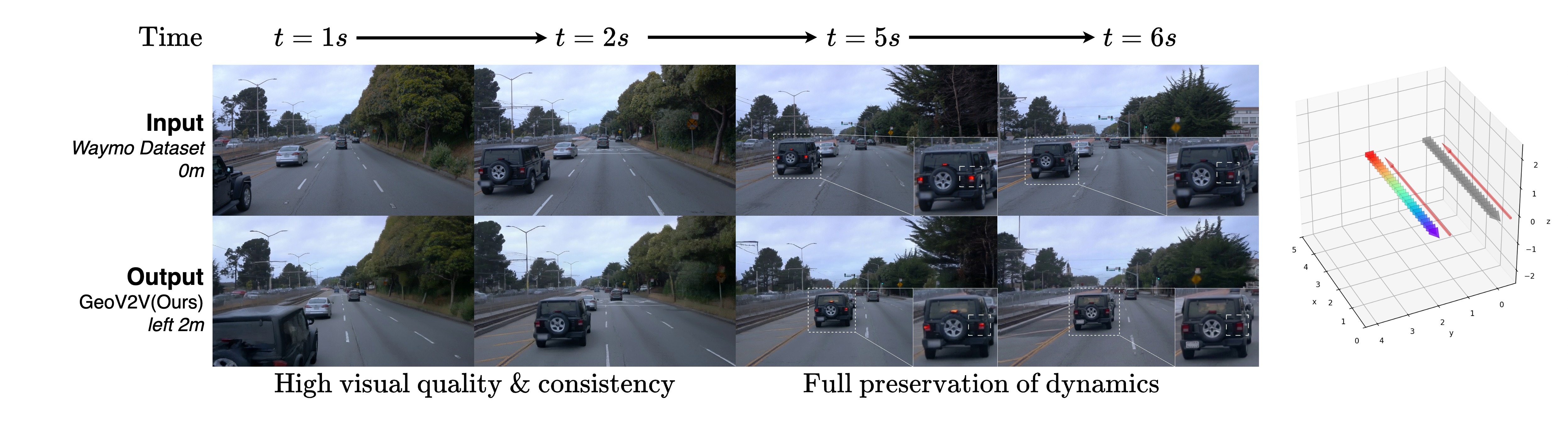}
    \caption{Our model achieves high visual quality and preserves dynamic lighting.}
\end{subfigure}

\caption{Overview and qualitative results of our method.}
\label{fig:overview_results}

\end{figure}

\begin{figure}[t]
    \centering
    \includegraphics[width=\linewidth]{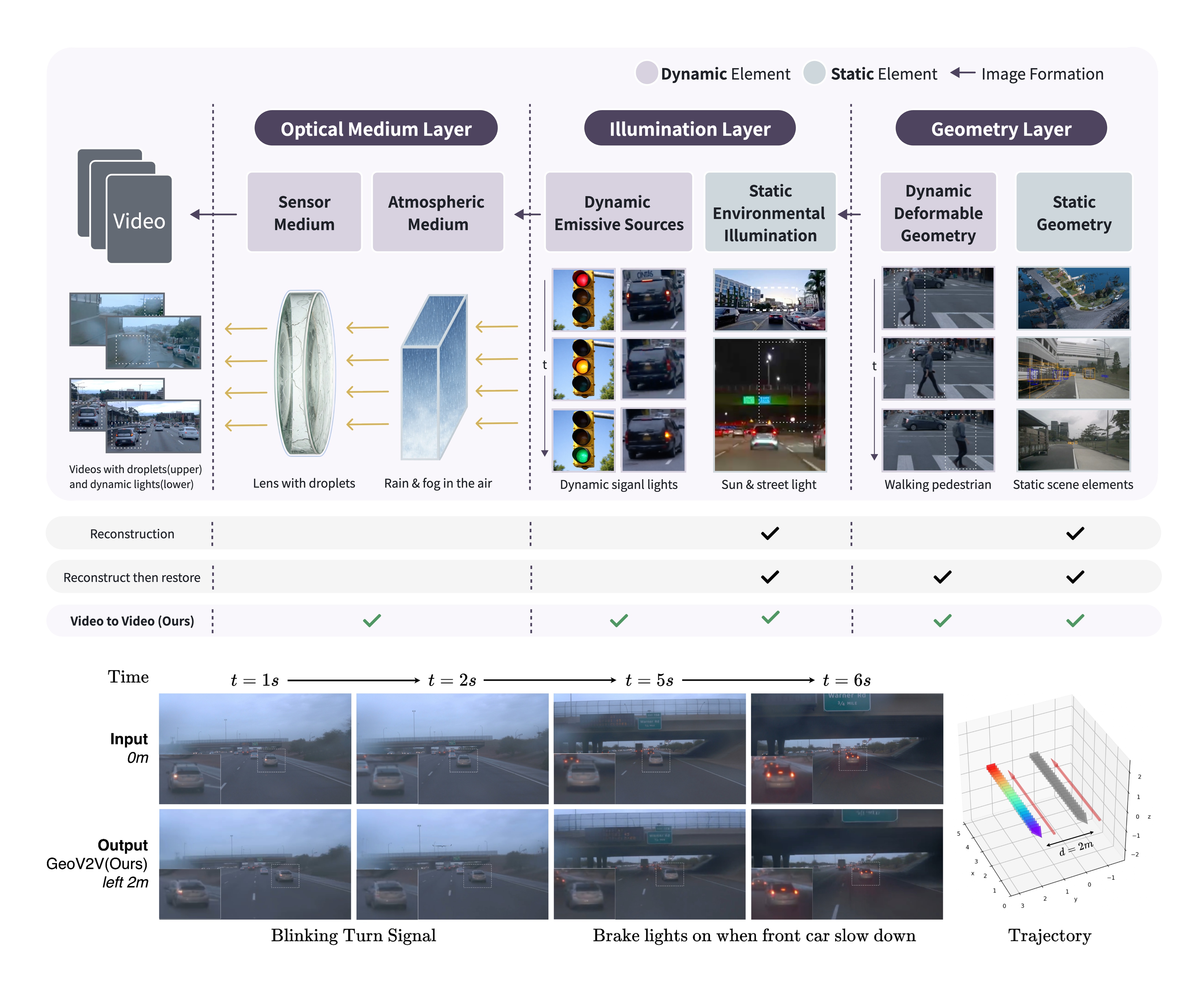}

    \caption{Motivation of our approach. Real-world videos arise from multiple factors including geometry, illumination dynamics, and optical media, making consistent synthesis difficult for existing methods (top). Our video-to-video model jointly preserves these factors and produces temporally consistent novel-view videos (bottom).}
    \label{fig:motiv}
\end{figure}

Existing approaches for solving this problem can be roughly categorized into reconstruction-based methods\cite{yang2023emernerf, yan2024street, zhou2024hugs} and reconstruct-then-restore methods\cite{ren2025cosmos,wang2024freevsgenerativeviewsynthesis, ren2025gen3c,wang2023drivedreamer,zhao2024drivedreamer4dworldmodelseffective}. Reconstruction-based methods, such as 3DGS\cite{kerbl20233d} and EmerNeRF\cite{yang2023emernerf}, rely on explicitly reconstructing the 3D scene through per-image optimization, then re-render the 3D scene from the new viewpoint. Although they often achieve high pixel accuracy (e.g., PSNR and SSIM) on interpolation tasks, they suffer severe quality degradation under extrapolation tasks\cite{Li2024XLD}. Reconstruct-then-restore methods, such as FreeVS\cite{wang2024freevsgenerativeviewsynthesis} and StreetCrafter\cite{2024StreetCrafter}, approach the problem by generating each frame conditioned on the colored LiDAR frame. They can synthesize visually coherent results within individual frames but often fail to maintain video-level visual consistency and hallucinate inconsistent 4D dynamics. These limitations arise from the reliance on explicitly modeling selected components of the scene. In real-world driving videos, however, visual appearance is jointly determined by more factors—including illumination dynamics and optical media—many of which are difficult to model explicitly (Fig.~\ref{fig:motiv}).

A fundamental obstacle to learning reliable cross-trajectory view synthesis directly lies in the lack of suitable training data. Existing autonomous driving datasets record videos along a single ego trajectory and do not provide synchronized observations from multiple parallel lanes within the same dynamic scene. Without such supervision, models must rely on pseudo-parallel views or misaligned camera signals, which introduce parallax mismatch and temporal inconsistencies. As a result, models struggle to learn the geometric transformations induced by lateral ego-motion, making stable lane-shift video synthesis difficult.

To address this limitation, we introduce \textbf{\dataset}, a synthetic dataset designed to provide explicit supervision for cross-lane view synthesis. Para4D captures synchronized videos from multiple parallel driving trajectories within the same dynamic scenes, enabling models to observe consistent 4D scene dynamics from shifted viewpoints. This structured supervision serves as a strong training prior for learning the geometry and motion patterns associated with lane shifting. Building on this dataset, we propose \textbf{\model}, a geometry-grounded video-to-video diffusion framework that integrates multi-modal geometric priors derived from LiDAR and depth estimation with full-video conditioning in the Wan 2.1~\cite{wan2025wanopenadvancedlargescale} backbone, enabling the generation of temporally stable and geometrically consistent driving videos along novel trajectories. Importantly, the end-to-end video generation framework also learns to preserve complex visual dynamics, such as changing illumination and other scene variations, without requiring explicit modeling of these factors, as shown in Fig.~\ref{fig:motiv}.

Our contributions are summarized as follows.

(1)%
We introduce \textbf{\dataset}, a synthetic dataset containing synchronized multi-lane driving videos captured within the same dynamic scenes. Para4D provides explicit supervision for cross-lane view synthesis and serves as a strong training prior for learning trajectory-shift transformations in 4D scenes.

(2) We propose \textbf{\model}, a video-to-video diffusion framework that integrates LiDAR and depth-based geometric priors with full-video conditioning to generate temporally coherent and geometrically consistent novel-view driving videos, without requiring any annotations like bounding boxes.

(3) Through extensive experiments, we compare reconstruction, reconstruct-then-restore, and our video-to-video paradigm for lane-shifting NVS, demonstrating the superior visual quality, geometric consistency, and dynamic preservation of the video-to-video paradigm. Results also show that synthetic-only training can achieve promising zero-shot transfer to real-world driving data.

%% file: sec/2_related.tex
\section{Related Work}
\textbf{Synthetic Driving Datasets.}
Synthetic datasets provide controllable and scalable environments for developing perception and generation models in autonomous driving. Early works leveraged game engines to amortize labeling cost: \textsc{SYNTHIA} introduced diverse urban scenes with pixel-accurate semantics for pretraining and domain transfer~\cite{Ros2016SYNTHIA}; Virtual KITTI cloned real KITTI sequences into a fully labeled virtual world with poses, depth, optical/scene flow, and segmentation—well matched to multi-view NVS supervision~\cite{Gaidon2016VKITTI}; and the GTA-V based dataset of Richter \etal supplied 24{,}966 densely labeled frames that became a cornerstone for synthetic-to-real segmentation~\cite{Richter2016PlayingForData}. Finally, \textsc{CARLA}~\cite{Dosovitskiy17CARLA} underpins much of this ecosystem as an open simulator producing multi-sensor, multi-view, pose-accurate streams across weather/lighting. Complementing these synthetic sources with an NVS-specific benchmark, XLD~\cite{Li2024XLD} targets cross-lane generalization by evaluating rendering from viewpoints that deviate meters off the training trajectory (front-only and multi-camera settings) across multiple weather/time conditions, thereby exposing gaps between current NVS methods and closed-loop AD simulation requirements. However, despite these advances, the community still lacks a highly dynamic, multi-camera autonomous-driving dataset specifically designed for studying view-consistent generation under lane-shifting and other large viewpoint shifting tasks.

\textbf{Reconstruction-based NVS Methods.}
Reconstruction-based novel view synthesis has greatly benefited from advancements in differentiable 3D rendering methods, exemplified by Neural Radiance Fields (NeRF)~\cite{mildenhall2020nerf} and 3D Gaussian Splatting (3DGS)~\cite{kerbl20233d}. These methods form the backbone of recent techniques for synthesizing novel views by learning representations of complex 3D scenes directly from images.
NeRF represents continuous volumetric scenes using multi-layer perceptrons, enabling the synthesis of novel views by predicting radiance and density at any 3D coordinate. Building on this foundation, many studies have extended its capabilities to reconstruct large-scale scenes~\cite{tancik2022block,turki2022mega,turki2023suds} and dynamic driving scenes~\cite{rematas2022urban,wu2023mars,xie2023s,yang2023unisim,tonderski2024neurad,wu2023mapnerf}.
3DGS represents the 3D environment using explicit Gaussian primitives, enabling faster training and rendering through the splat-based rasterization method. Originally designed for static and bounded scenes, 3DGS has been extended to handle general dynamic scene reconstruction~\cite{wu20234dgaussians,yang2023deformable3dgs,yang2023gs4d}, and dynamic urban driving scenes~\cite{zhou2024drivinggaussian,yan2024street,fischer2024dynamic,huang2024s3gaussian}.
To enhance the modeling of the complex and dynamic environments, some works separate static and dynamic elements and incorporate semantic information~\cite{fischer2024dynamic, yan2024street,zhou2024drivinggaussian}.
Due to the lack of sufficient data needed for novel view synthesis, reconstruction methods face significant challenges in accurately rendering views that are merely seen.

\textbf{Generation-based NVS Methods.}
Novel view synthesis through image generation has greatly benefited from the advancements in diffusion models \cite{ho2020denoising,song2020denoising,rombach2022high,blattmann2023stable,wan2025wanopenadvancedlargescale}. Various types of information are encoded by diffusion models to help generate new images as extra guidance for the reconstruction process, e.g., a reference image and the target camera pose embedded as a text embedding \cite{liu2023zero,zeronvs}, volume-rendered feature images \cite{chan2023generative}, images rendered by the reconstruction model \cite{wu2024reconfusion}, extra perspective view images \cite{melas2023realfusion}.

Recently, there has been growing work in applying diffusion models to driving scenarios \cite{zhao2024drivedreamer4dworldmodelseffective,yu2024sgd,wang2024freevsgenerativeviewsynthesis}. DriveDreamer4D \cite{zhao2024drivedreamer4dworldmodelseffective} proposed a novel data augmentation module that can be integrated with other methods to enhance NVS performance. FreeVS \cite{wang2024freevsgenerativeviewsynthesis} proposed a full generative model to generate target images from pseudo-images, which are generated by projecting colored LiDAR points onto specific positions.

Different from these methods, our approach leverages a geometry-guided video-to-video diffusion model trained on large-scale synthetic data to achieve strong visual quality and spatiotemporal consistency.

%% file: sec/3_dataset.tex
\section{\dataset\ Synthetic Dataset}

Paired multi-lane video sequences captured from synchronized parallel trajectories in dynamic scenes are required for training lane-shifting NVS models. However, collecting such data in real-world driving is impractical, and existing synthetic datasets mainly focus on static scenes \cite{Li2024XLD}.

\begin{figure}[t]
    \centering
    \includegraphics[width=1\linewidth]{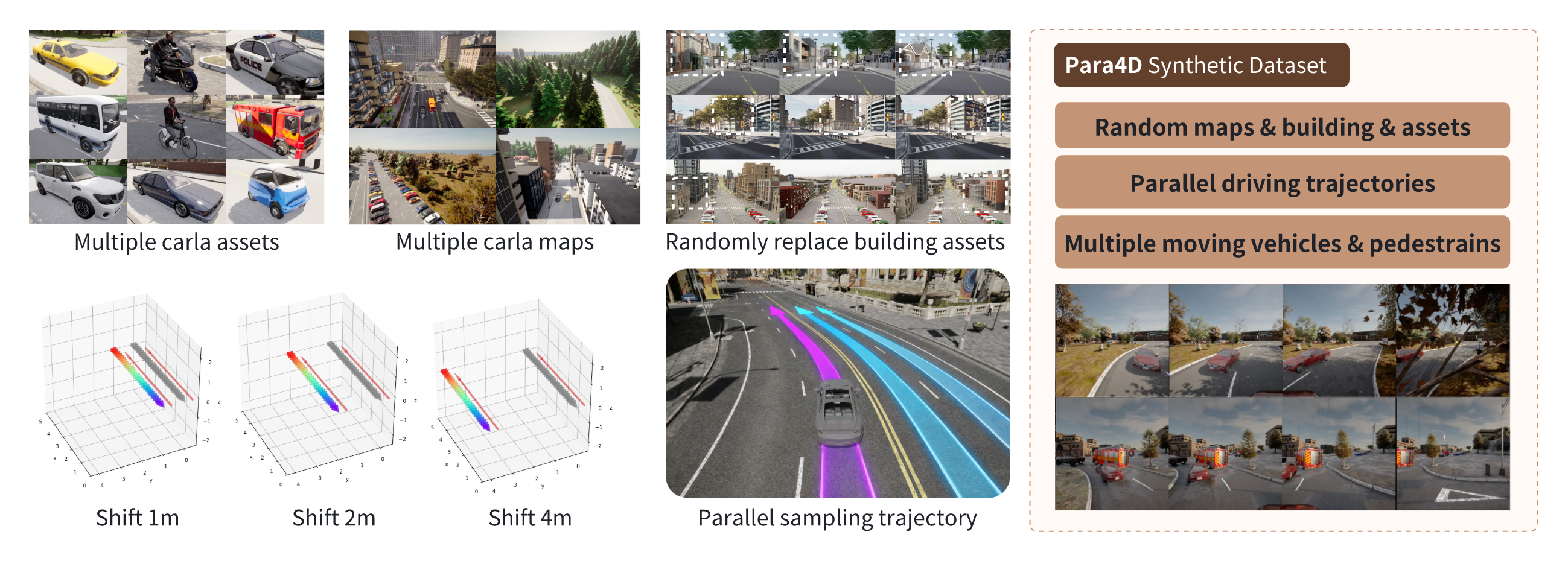}
    \caption{Overview of the Para4D dataset. We generate multi-lane synchronized driving videos in the CARLA simulator using diverse maps, randomized assets, and dynamic traffic agents. Parallel camera trajectories are placed across lanes to render synchronized lane-shifted videos for training.}
    \label{fig:Para4D}
\end{figure}

To address this limitation, we introduce \dataset, a synthetic dataset built using the CARLA simulator \cite{Dosovitskiy17CARLA}. It enables precise control over camera extrinsics, perfect temporal synchronization, and scalable generation of dynamic urban driving scenes with diverse layouts, weather conditions, and traffic agents. We build our data generation pipeline on the CARLA simulator to construct diverse urban environments by randomly sampling maps and replacing building assets. Dynamic agents, including vehicles and pedestrians, are instantiated and controlled by CARLA’s traffic system to produce realistic interactions. Multiple camera sets are placed on parallel lanes and move synchronously with the ego vehicle, with lateral offsets of up to 4 meters. This setup enables rendering synchronized multi-lane driving videos that share the same dynamic scene while providing different viewpoints, allowing reliable training of lane-shifting NVS models.

%% file: sec/4_method.tex
\section{Method}

Given a driving scene video containing rich dynamic content, our goal is to synthesize novel-view videos along a specified camera trajectory while ensuring both temporal coherence and geometric consistency.

To achieve this, we rely on two inputs: 3D priors and the full source video. We construct the 3D priors from LiDAR and image-derived depth, providing explicit geometric guidance without requiring human annotations such as 3D boxes for moving vehicles. The full source video from the original viewpoint, rather than isolated frames, allows the model to preserve appearance consistency, fine-grained scene details, and temporal dynamics. Our video diffusion model, GeoV2V, then jointly conditions on both inputs to generate novel-view videos that are temporally stable and geometrically consistent across diverse driving scenarios.

\subsection{3D Priors Construction}
\label{sec:3d_prior}

\begin{figure}[t]
    \centering
    \includegraphics[width=0.75\linewidth]{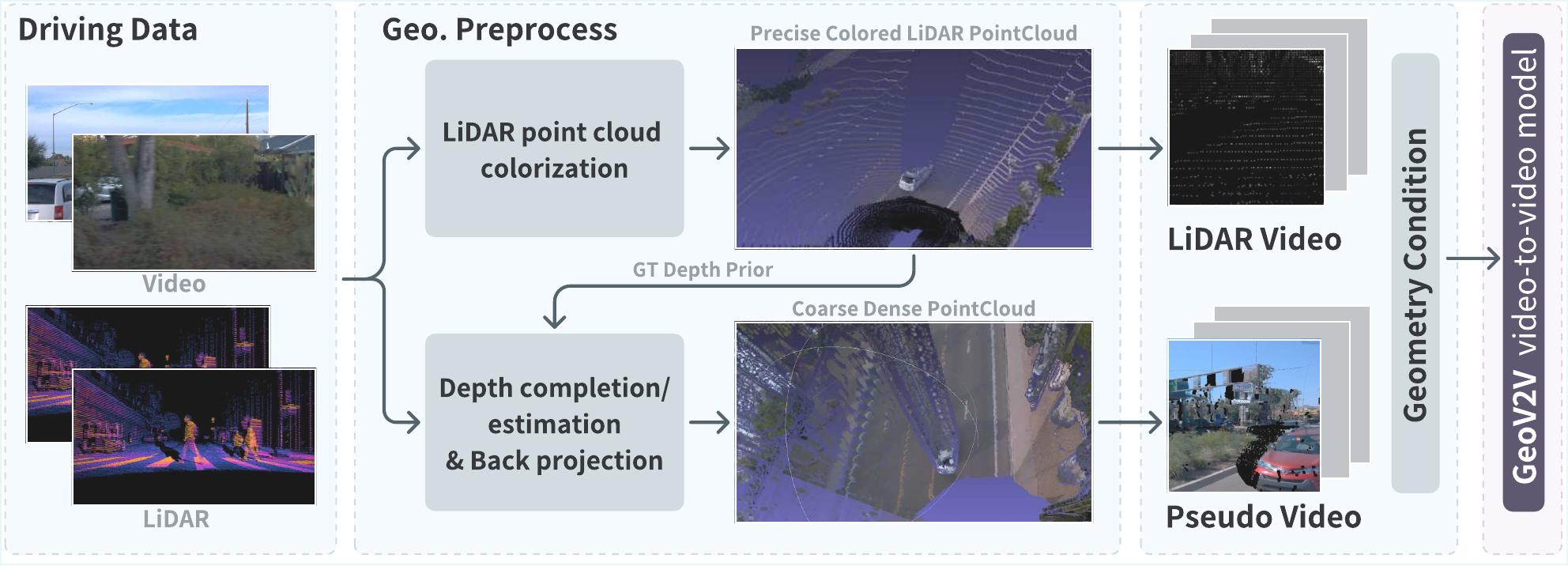}
    \caption{Geometry prior preprocessing.}
    \label{fig:depth_estimation}
\end{figure}

\textbf{Global 3D Representations.}
As shown in Fig.~\ref{fig:depth_estimation}, to capture the complete 3D structure of each individual frame, we construct per-frame global 3D representations by fusing LiDAR-colored point clouds with depth-completed reprojected point clouds. Notably, we only establish geometric representations at the single-frame level, without introducing explicit temporal modeling or inter-frame constraints, so as to avoid weakening the temporal consistency and dynamics modeled by the subsequent video diffusion framework. While LiDAR point clouds provide highly accurate geometric measurements, they are inherently sparse and often fail to densely cover the scene. To mitigate this limitation, we incorporate dense reprojected point clouds derived from depth estimation\cite{chen2025videodepthanythingconsistent} and completion\cite{liang2025distillingmonocularfoundationmodel}, which complement the sparse LiDAR data and provide richer geometric cues.

\textbf{Rendering Novel-View Priors.}
\label{sec:method}
Given the constructed multi-modal 3D representations for each frame, we render them into target viewpoints to obtain novel-view priors. LiDAR-colored points and depth-completed reprojected points are projected separately to preserve their complementary properties: LiDAR points provide accurate yet sparse geometry, while dense reprojected points offer broader coverage and finer structural details.

Even after densification, the projected point cloud still exhibits mesh-like void artifacts in novel views due to discretization, depth discontinuities, and viewpoint-dependent occlusions. These holes vary with disparity and camera motion, introducing temporal inconsistency and destabilizing downstream generation. Naive hole-filling strategies such as nearest-neighbor interpolation or simple inpainting tend to violate perspective geometry and may incorrectly fill originally occluded regions, thereby misleading the diffusion-based synthesis model.

To address this issue, we adopt a depth-aware morphological filling strategy. Let $M$ denote the binary validity mask. We first obtain candidate fillable regions via morphological closing:

\begin{equation}
\Omega = \big((M \oplus \mathcal{S}) \ominus \mathcal{S}\big) \setminus M.
\end{equation}

For each $u \in \Omega$, we identify its nearest valid pixel
\begin{equation}
u_n = \arg\min_{v \in M} |u - v|_2,
\end{equation}

and define the depth-consistent neighborhood
\begin{equation}
\mathcal{N}_d(u) = \{ v \in M \mid |u - v|_2 < r,\; |D(v) - D(u_n)| < \tau_d \}.
\end{equation}

The filled value is then computed by distance-weighted averaging:
\begin{equation}
\hat{I}(u) =
\frac{\sum_{v \in \mathcal{N}_d(u)} w(u,v) I(v)}
{\sum_{v \in \mathcal{N}_d(u)} w(u,v)},
\quad
w(u,v)=\exp!\left(-\frac{|u-v|_2^2}{\sigma^2}\right),
\end{equation}
where $\mathcal{N}_d(u)$ denotes neighbors satisfying the depth constraint, and $\sigma$ controls spatial weighting.

While such a depth-constrained morphological strategy may not be universally effective in all scenarios, it avoids introducing excessive errors and alleviates hallucinations in the downstream generative model, thus facilitating more reliable geometric priors for diffusion-based novel-view synthesis.

\begin{figure}[t]
    \centering
    \includegraphics[width=1\linewidth]{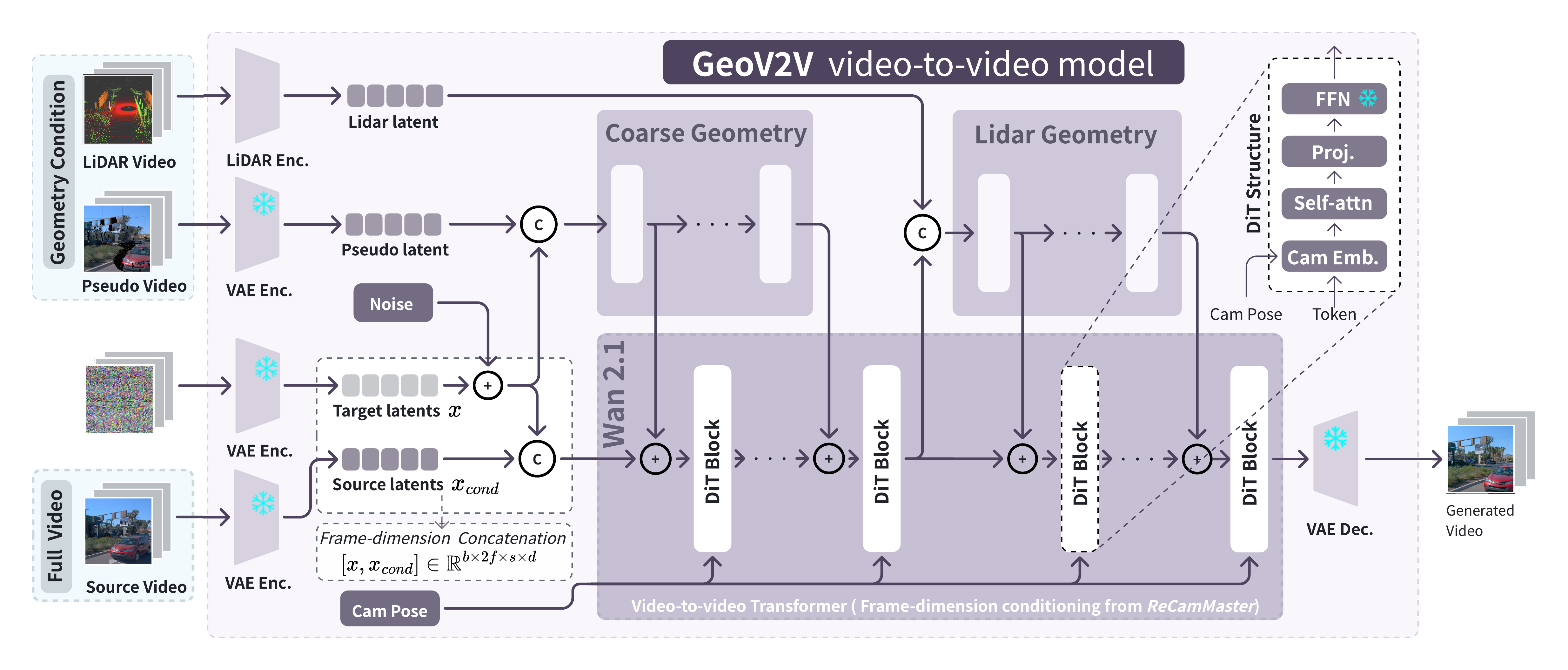}
    \caption{Architecture of GeoV2V, a geometry-guided video-to-video diffusion model. The model encodes source, pseudo, target (noisy), and LiDAR videos into latent representations. Through frame-wise conditioning with geometry signals and camera-pose embeddings, the DiT backbone performs geometry-consistent video-to-video synthesis for novel-view generation.}
    \label{fig:model_structure}
\end{figure}
\subsection{Geometry-aware Video Diffusion Model}
\label{sec:dual_priors}
\textbf{Model overview.}
An overall framework of \model\ is presented in Fig.~\ref{fig:model_structure}. Our framework builds upon the Wan-2.1 video diffusion \cite{wan2025wanopenadvancedlargescale} backbone, a large-scale transformer-based architecture capable of capturing long-range temporal dependencies and fine-grained spatial details. Given the input video $\mathbf{V}_{\text{src}} = \{I_t\}_{t=1}^T$ and the constructed geometry priors $\mathbf{P}_{\text{geo}} = \{D_t, S_t\}_{t=1}^T$, where $D_t$ denotes the dense depth-completed reprojected points, and $S_t$ denotes the separate projection for LiDAR, the model learns to generate the corresponding video $\mathbf{V}_{\text{tgt}}$ observed from the novel trajectory.

\textbf{Source video conditioning.}
In novel view generation, the source-view video contains the richest scene details and temporal dynamics. However, there is no explicit spatial correspondence between the source view and the target view during denoising.

To better exploit the source information, we follow the conditioning scheme of Bai et al.\cite{bai2025recammastercameracontrolledgenerativerendering} and introduce conditional embeddings in both temporal and spatial dimensions. For the temporal dimension, we leverage the Rotary Positional Embedding (RoPE) used in video diffusion models and concatenate the source-view latents with the target-view noisy latents along the frame dimension. This design keeps temporally aligned source-target frames at a fixed relative offset in the concatenated sequence, without introducing any additional attention mechanism. For spatial conditioning, we encode per-frame camera extrinsics and inject the resulting embeddings into the corresponding frame latents, enabling the model to incorporate view-dependent geometric cues during denoising.

\textbf{Geometry injecting.}
Since raw camera extrinsics alone cannot provide the diffusion model with accurate geometric understanding, we further inject the two geometric priors obtained in Sec.~\ref{sec:3d_prior} into the denoising process.

Considering the complementary characteristics of the two priors, we adopt a hierarchical injection strategy, where the first 15 layers and the last 15 layers of the denoising network are respectively conditioned on the two priors. This design is motivated by AC3D\cite{bahmani2025ac3d}, which suggests that early DiT layers mainly determine coarse camera viewpoints. The design of this 15-layer split is to ensure that the control flows of the two priors do not interfere with each other: the dense full-depth reprojection prior is introduced first in the initial 15 layers to quickly determine the overall structural information of the scene, laying a solid geometric foundation for subsequent generation; then the sparse LiDAR reprojection prior is used in the last 15 layers to supervise and correct the fine-grained geometric details, further improving the 3D consistency of the generated results. Due to the inherent sparsity of LiDAR priors, we encode them separately using a dedicated ConvNeXt\cite{liu2022convnet} encoder to better retain fine-grained geometric information and avoid information loss caused by sparsity.

Notably, both branches are implemented as lightweight ControlNets (61M parameters each), much smaller than the 1.3B DiT backbone. This avoids excessive geometric constraints and preserves rich appearance details from the source video, balancing 3D consistency and fidelity.

\textbf{Training strategy.}
To ensure strong generalization capability across diverse scenes, we freeze the feed-forward network (FFN) layers of the WAN-DiT backbone and fine-tune only the self-attention blocks together with the camera embedding modules.
This strategy allows the model to adapt its spatial-temporal reasoning to novel view configurations while maintaining stable feature representations learned from large-scale pretraining.

Since the LiDAR patterns vary significantly across datasets, the LiDAR-injecting branch is trained separately after the rest of the network has converged.
This two-stage training scheme prevents overfitting to dataset-specific LiDAR distributions and improves the model’s transferability to unseen data.

%% file: sec/5_experiment.tex
\section{Experiments}
\label{sec:exp}

\subsection{Experiment Setup}
\label{sec:experiment_setup}

\subsubsection{Datasets and Metrics}
We evaluated our method on three datasets of varying realism and complexity: \textit{Waymo Open Dataset}~\cite{sun2020scalability}, \textit{nuScenes}~\cite{caesar2020nuscenes}, and our \dataset.
The first two are large-scale real-world autonomous driving datasets that provide multi-camera videos and LiDAR point clouds under diverse urban conditions.
However, since neither contains paired trajectories with lateral shifts, we use them for \textit{off-record novel-view generation} evaluation.
In this setting, our model is tested on new camera trajectories parallel to the original lane, and performance is quantified using the Fréchet Inception Distance (FID), which reflects perceptual realism and cross-view consistency.

Para4D contains synchronized videos captured from multiple cameras on parallel lanes.
For each scene, we also record LiDAR point clouds and accurate camera parameters, providing ground-truth correspondences between trajectories.
This enables us to compute pixel-level quantitative metrics including PSNR, SSIM, and LPIPS between generated and reference videos, evaluating visual fidelity and geometric consistency.

\subsubsection{Baseline Methods}
We compare our approach against several representative baselines that cover both geometry-based and video-based paradigms: \textbf{StreetGaussian}~\cite{yan2024street}: a 3D Gaussian Splatting method tailored for street-scale reconstruction, requiring bounding boxes for dynamic vehicles; \textbf{OmniRe}~\cite{chen2024omnire}: a method that improves StreetGaussian by introducing deformable pedestrian modeling; \textbf{EmerNeRF}~\cite{yang2023emernerf}: a neural-field framework that jointly models static structures, dynamic motion, and induced flow fields; \textbf{FreeVS}~\cite{wang2024freevsgenerativeviewsynthesis}: a video synthesis framework that uses a diffusion model to restore colored LiDAR point clouds; \textbf{StreetCrafter}~\cite{2024StreetCrafter}: a controllable video diffusion model that conditions on pixel-level LiDAR renderings to synthesize novel views while preserving camera control; \textbf{DiST-4D}~\cite{2025DiST}: a disentangled spatiotemporal diffusion framework that uses metric depth as the core geometric representation to jointly support temporal prediction and spatial novel view synthesis. In our experiments, DiST-4D adopts the same depth as our method.

As shown in Table~\ref{tab:exp_real_interp}, compared with most methods that explicitly model dynamics, our approach does not require labor-intensive, manually annotated bounding boxes and thus exhibits stronger generalizability.

\subsection{Comparison Results}
The quantitative comparisons are presented in Tables~\ref{tab:exp_real_interp}-\ref{tab:exp_carla_shift}. Tables~\ref{tab:exp_real_interp} and \ref{tab:exp_real_shift} show results on real-world datasets, while Table~\ref{tab:exp_carla_shift} reports results on our Para4D dataset.

\begin{figure*}[t]
    \centering
    \includegraphics[width=0.97\linewidth]{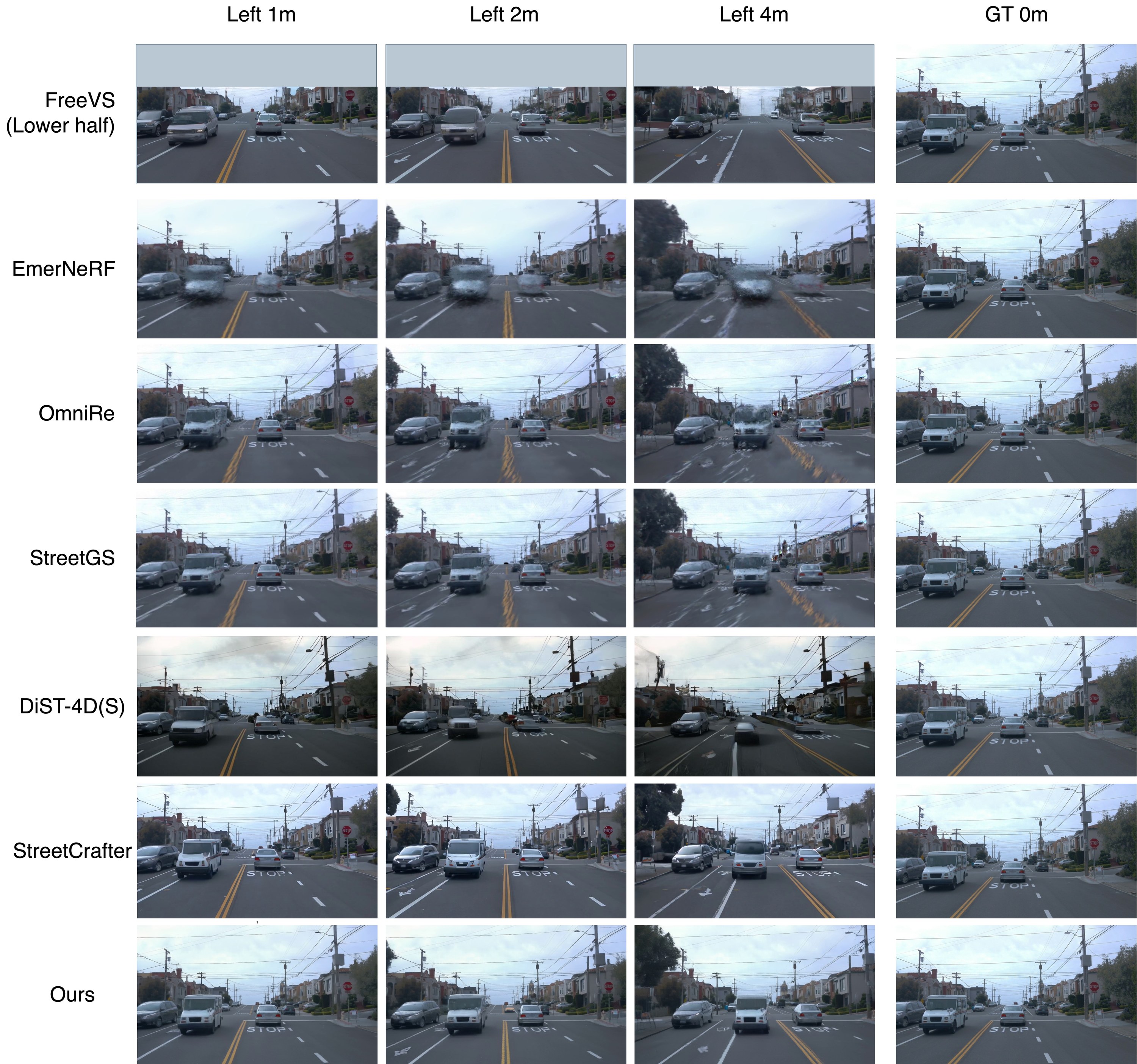}
    \caption{
        Qualitative new-view synthesis results on the Waymo dataset, where FreeVS\cite{wang2024freevsgenerativeviewsynthesis} is evaluated using the original cropping strategy in their official implementation, resulting in synthesis restricted to only the lower portion of the scene.
    }
    \label{fig:waymo_nvs_1}
\end{figure*}
\begin{figure}
    \centering
    \includegraphics[width=1\linewidth]{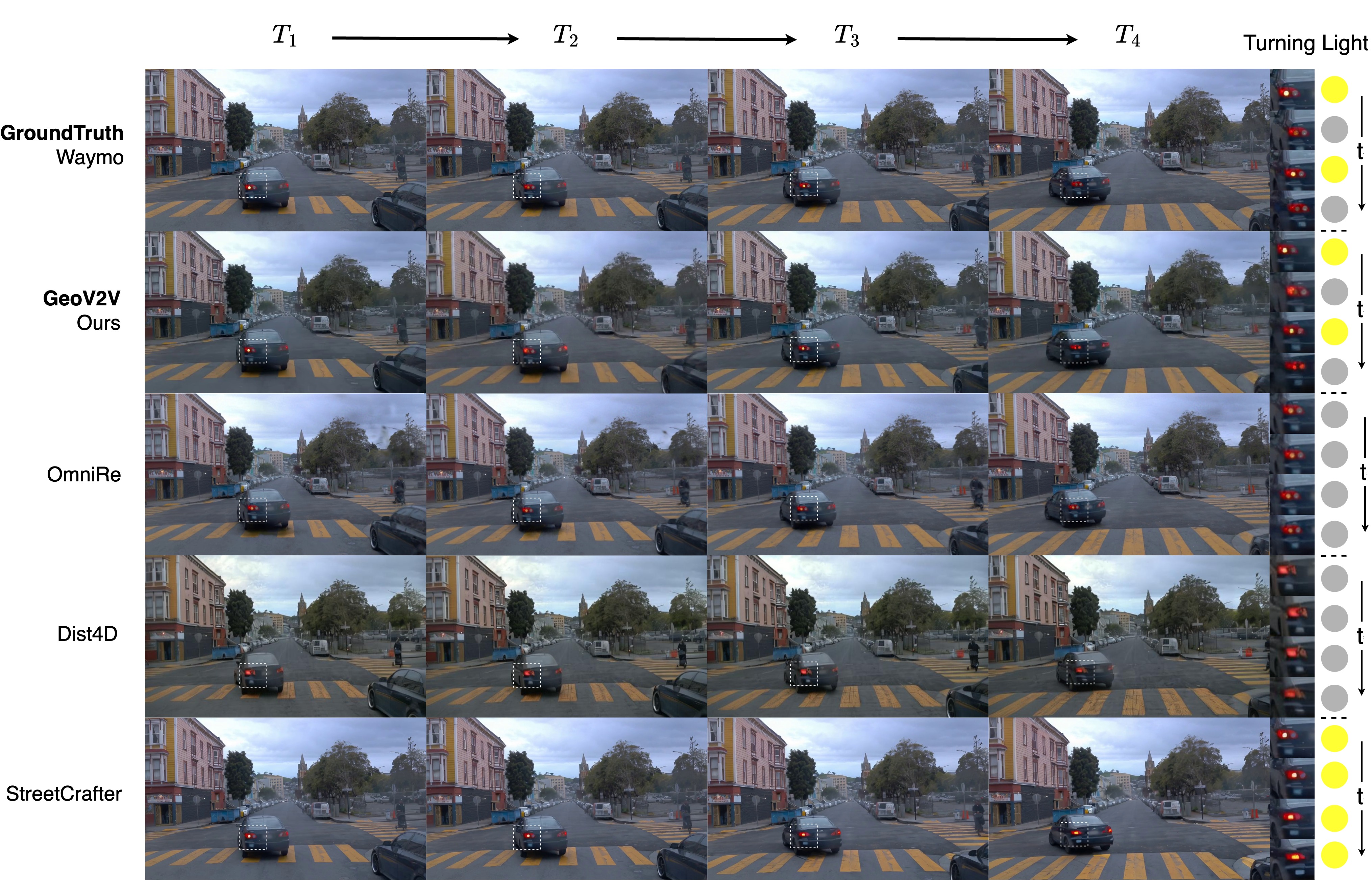}
    \caption{Comparison of temporal dynamic lighting preservation. GeoV2V faithfully preserves the vehicle's turn signal across time steps, while existing methods fail to maintain consistent dynamic lighting patterns.}
    \label{fig:blinking_comparison}
\end{figure}

\begin{table}[h!]
\centering
\caption{Comparison on real-world datasets (interpolation).}
                    \label{tab:exp_real_interp}
\begin{tabular}{|l|l|ccc|ccc|}
\toprule
\multirow{2}{*}{Method}  &\multirow{2}{*}{Annotation}&
\multicolumn{3}{c|}{Waymo Static} &
\multicolumn{3}{c|}{Waymo Dynamic}  \\
 && PSNR & SSIM & LPIPS & PSNR & SSIM & LPIPS \\
\midrule
FreeVS
 &Bounding-box& 22.77& 0.670& 0.350
& 21.78& 0.596& 0.362
\\
 OmniRe &Bounding-box& 24.10& 0.708
& 0.252
& 23.73& 0.683& 0.237\\
 EmerNeRF
 &Bounding-box& 24.92& 0.701&
0.245& \textbf{23.91}& \textbf{0.698}&
0.304\\
StreetGS
 &Bounding-box& 24.37 & 0.714 & 0.239
& 23.54& 0.696& 0.256
\\
 StreetCrafter-V &Bounding-box&19.46 &0.611 &0.353  &18.92 &0.597 &0.353\\
 DiST-4D &\textbf{None}&18.52 &0.566 &0.378 &18.15 &0.563 &0.373\\

Ours
 &\textbf{None}& \textbf{25.46}& \textbf{0.716}& \textbf{0.132}& 23.42 & 0.692 & \textbf{0.149}\\
\bottomrule
\end{tabular}
\end{table}

\textbf{Interpolation on original trajectories.}
Following previous 3DGS works, we first evaluate our method on the original camera trajectories using an interpolation setting. We adopt a $\{1\!-\!0\!-\!1\!-\!0\!-\!1\}$ temporal sampling scheme, where a video sequence $\{I_t\}_{t=1}^{T}$ is split into an input set $\mathcal{I}_{\mathrm{in}}=\{\, I_t \mid t \bmod 2 = 1 \,\}$ and a held-out set $\mathcal{I}_{\mathrm{eval}}=\{\, I_t \mid t \bmod 2 = 0 \,\}$. Frames in $\mathcal{I}_{\mathrm{in}}$ are used for training or conditioning the model, while the skipped frames in $\mathcal{I}_{\mathrm{eval}}$ are reconstructed and used exclusively for testing and quantitative evaluation\cite{wang2024freevsgenerativeviewsynthesis}. Metrics including PSNR, SSIM, and LPIPS are computed on these held-out frames. As shown in Table~\ref{tab:exp_real_interp}, our model achieves the best overall performance across all datasets and metrics, setting a new state-of-the-art baseline for trajectory interpolation.

\begin{table}[h!]
\centering
\caption{Comparison on real-world datasets (lane shifting, FID \& FVD).}
\label{tab:exp_real_shift}
\begin{tabular}{l|ccc|ccc}\toprule
 & \multicolumn{3}{c}{FID}& \multicolumn{3}{c}{FVD}\\

&
$1$~m &
$2$~m &
$4$~m &
$1$~m &
$2$~m &
$4$~m \\\midrule

FreeVS
&39.93 & 47.35 & 51.89 & 335.32 & 373.57 &445.36\\
 OmniRe& 26.46& 34.59& 45.63 & 223.29& 334.43&601.84\\
 EmerNeRF& 45.74& 55.25& 73.23 & 346.90& 445.55&661.26\\
StreetGS
& 24.35& 30.52& 46.96 & \underline{212.56} & 331.28&552.19\\
 StreetCrafter& \underline{18.69}& \textbf{22.56}& \textbf{28.53}& 217.30 & \underline{272.85}&\underline{369.64}\\
 DiST-4D& 22.47& 28.16& 38.13& 246.38 & 369.76&552.62\\

Ours
& \textbf{17.53}& \underline{23.59}& \underline{32.58} & \textbf{159.00}& \textbf{232.06}&\textbf{328.69}\\ \bottomrule

\end{tabular}
\end{table}

\textbf{Lane-shifting on real-world datasets.}
Most existing 3DGS methods do not evaluate lane-shifting, while generation-based approaches such as FreeVS\cite{wang2024freevsgenerativeviewsynthesis}, Recondreamer\cite{ni2025recondreamer}, and FreeSim\cite{fan2025freesim} typically report FID. Following this convention, we evaluate both FID and FVD, where FID measures frame-level visual quality and FVD evaluates video temporal consistency. As shown in Table~\ref{tab:exp_real_shift}, our model achieves strong performance on FID and the best score on FVD, producing temporally coherent and visually realistic lane-shifted videos. Notably, the model shows promising transfer to real-world scenes even when trained solely on synthetic data, without additional fine-tuning.

\begin{table}[h!]
\centering
\caption{Comparison on the Para4D dataset.}
\label{tab:exp_carla_shift}
\resizebox{\columnwidth}{!}{
\begin{tabular}{l|cccc|cccc|cccc}
\toprule
\multirow{2}{*}{Method}  &
\multicolumn{4}{c|}{$1$~m} &
\multicolumn{4}{c|}{$2$~m} &
\multicolumn{4}{c}{$4$~m} \\
 & PSNR & SSIM & LPIPS & FID & PSNR & SSIM & LPIPS & FID & PSNR & SSIM & LPIPS & FID  \\
\midrule
FreeVS*
 & 16.35& 0.512& 0.568& 118.72& 12.58& 0.418& 0.689& 152.43& 12.32& 0.375& 0.702& 198.25\\
 EmerNeRF & 22.56& 0.701& 0.232& 35.56& 21.22& 0.695& 0.386& 65.04& 19.10& 0.614& 0.502&83.29\\
StreetGS
 & 24.34& 0.714 & 0.172 & 27.75
& 23.00& 0.696 & 0.181& 37.39
& 19.85 & 0.590& 0.291& 89.67 \\
 StreetCrafter &19.35 &0.578 &0.379 &46.02 &17.05 &0.527 &0.514 &64.25 & 16.65& 0.503& 0.601&96.02\\
 DiST-4D &19.52 &0.602 &0.368 &47.35 &16.96 &0.503 &0.622 &74.65 &14.21 & 0.459&0.659 &105.62\\
Ours
 & \textbf{25.18}& \textbf{0.738}& \textbf{0.040}& \textbf{17.53}& \textbf{23.88}& \textbf{0.697}& \textbf{0.051}& \textbf{25.25}& \textbf{21.31}& \textbf{0.608}& \textbf{0.080}& \textbf{41.17}\\
\bottomrule
\end{tabular}
}
\end{table}

\textbf{Lane-shifting on synthetic datasets.}
The Para4D dataset provides perfectly synchronized videos across multiple parallel lanes within the same dynamic scene, making it possible to evaluate both pixel-level accuracy and perceptual realism. We therefore report PSNR, SSIM, LPIPS, and FID scores. As summarized in Table~\ref{tab:exp_carla_shift}, our model achieves consistently higher PSNR and SSIM, indicating stronger geometric alignment, while also outperforming all baselines on perceptual metrics. The combination of superior pixel fidelity and perceptual realism highlights the model's ability to maintain both accurate 3D structure and natural temporal motion under large lateral shifts.

\textbf{Preservation of complex visual dynamics.}
Across multiple qualitative examples, we observe that GeoV2V preserves complex visual dynamics that are difficult to model explicitly. As illustrated in Fig.~\ref{fig:overview_results} and Fig.~\ref{fig:blinking_comparison}, the model faithfully maintains temporal lighting patterns such as blinking turn signals and brake lights across frames. Moreover, Fig.~\ref{fig:motiv} shows that our approach also preserves challenging optical effects, including lens droplets and other imaging artifacts, while maintaining consistent scene dynamics under viewpoint changes. These results suggest that the end-to-end video generation framework can implicitly capture complex video formation factors without requiring explicit modeling.

\subsection{Ablation Studies}

\begin{table}[h!]
    \centering
    \caption{Ablation studies on different modules.}
    \label{tab:exp_ablation}
    \begin{tabular}{c|ccc}\toprule
         Priors&  \multicolumn{3}{c}{$2$~m}\\
        & PSNR& SSIM& LPIPS  \\\midrule
full priors & 23.88 & 0.697 & 0.051\\
w/o video&  23.28 & 0.695 & 0.052 \\
w/o geometry (dense)& 17.18  & 0.504  & 0.124 \\
w/o geometry (sparse)& 22.86 & 0.668 & 0.058 \\
w/o geometry (both)& 16.45 & 0.490 & 0.128 \\
w/o Para4D & 20.92 & 0.593 & 0.102 \\\midrule
\end{tabular}
\end{table}

As shown in \cref{tab:exp_ablation}, we observe the importance of these three main components.
\begin{figure*}
    \centering
    \includegraphics[width=1\linewidth]{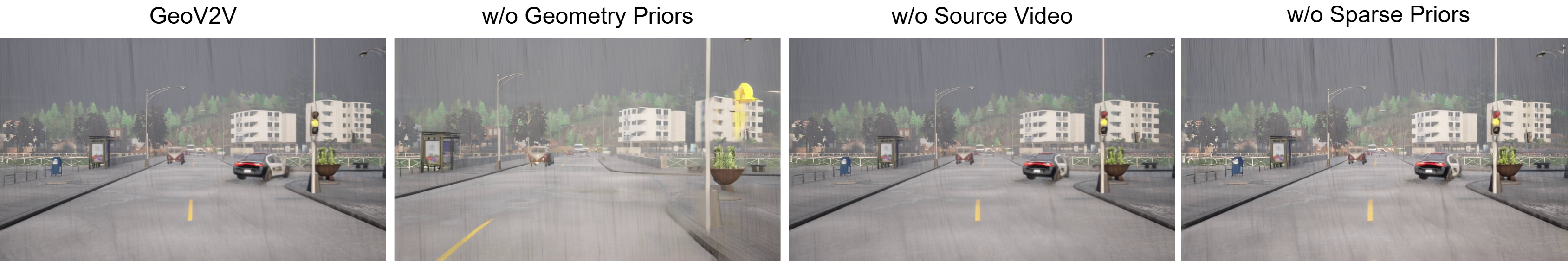}
    \caption{Qualitative results of ablation study.}
    \label{fig:carla_abl}
\end{figure*}

\textbf{Video input.} The input from the original view ensures the consistency of the generated content. As shown in Fig.~\ref{fig:carla_abl}, when the video input is removed, the generated vehicles become distorted due to depth errors in the geometric prior used for rendering, although the quantitative metrics do not drop significantly because the video input mainly acts on fine details where geometry is difficult to determine.

\textbf{Geometry priors.} Geometry priors ensure spatial alignment between the generated frames and the true 3D structure of the scene. They constrain the video diffusion process with depth-aware reprojection and LiDAR-informed priors. Without these priors, as shown in Table~\ref{tab:exp_ablation}, all metrics decrease to varying degrees. In particular, after removing the dense geometry prior, as illustrated in Fig.~\ref{fig:carla_abl}, the model fails to respond correctly to geometric offsets.

\textbf{Multi-lane Para4D training.} Existing real-world datasets lack synchronized parallel trajectories, making direct supervision for lane shifting infeasible. Without Para4D, a common alternative is to approximate lane shifts using misaligned side views, which introduces parallax mismatch and temporal inconsistency. To quantify the impact of this supervision structure, we train GeoV2V using either paired synchronized Para4D or misaligned side views generated from Para4D and Waymo. As shown in Table~\ref{tab:exp_ablation}, multi-lane supervision yields substantially higher PSNR/SSIM and lower LPIPS on Para4D, demonstrating that synchronized multi-lane supervision provides a strong training prior necessary for learning stable cross-lane transformations.

%% file: sec/6_conclusion.tex
\section{Conclusion}

This paper introduces GeoV2V, a geometry-grounded video-to-video diffusion framework for novel view synthesis in dynamic driving scenes without explicit dynamic modeling. We also present Para4D, a synthetic dataset providing synchronized multi-lane driving videos within the same dynamic scenes, enabling explicit supervision for cross-lane view synthesis. Experiments show that GeoV2V achieves strong visual quality and geometric consistency while preserving complex visual dynamics such as changing lights and optical artifacts, even without explicitly modeling them, and shows promising transfer to real-world driving scenes when trained only on synthetic data.

An important direction for future work is developing evaluation metrics that better capture the preservation of complex visual dynamics. Standard pixel-level metrics such as PSNR or SSIM are largely insensitive to effects like dynamic lighting or optical artifacts, making them less suitable for assessing these aspects. Currently, such phenomena are mainly evaluated qualitatively. Designing metrics that measure these factors more effectively would be valuable, especially for lane-shifting simulation, where effects such as lens contamination or changing traffic signals often represent critical corner cases for autonomous driving systems.

%% file: main.bbl
\begin{thebibliography}{10}
\providecommand{\url}[1]{\texttt{#1}}
\providecommand{\urlprefix}{URL }
\providecommand{\doi}[1]{https://doi.org/#1}

\bibitem{bahmani2025ac3d}
Bahmani, S., Skorokhodov, I., Qian, G., Siarohin, A., Menapace, W.,
  Tagliasacchi, A., Lindell, D.B., Tulyakov, S.: Ac3d: Analyzing and improving
  3d camera control in video diffusion transformers. In: Proceedings of the
  Computer Vision and Pattern Recognition Conference. pp. 22875--22889 (2025)

\bibitem{bai2025recammastercameracontrolledgenerativerendering}
Bai, J., Xia, M., Fu, X., Wang, X., Mu, L., Cao, J., Liu, Z., Hu, H., Bai, X.,
  Wan, P., et~al.: Recammaster: Camera-controlled generative rendering from a
  single video. arXiv preprint arXiv:2503.11647  (2025)

\bibitem{blattmann2023stable}
Blattmann, A., Dockhorn, T., Kulal, S., Mendelevitch, D., Kilian, M., Lorenz,
  D., Levi, Y., English, Z., Voleti, V., Letts, A., et~al.: Stable video
  diffusion: Scaling latent video diffusion models to large datasets. arXiv
  preprint arXiv:2311.15127  (2023)

\bibitem{caesar2020nuscenes}
Caesar, H., Bankiti, V., Lang, A.H., Vora, S., Liong, V.E., Xu, Q., Krishnan,
  A., Pan, Y., Baldan, G., Beijbom, O.: nuscenes: A multimodal dataset for
  autonomous driving. In: Proceedings of the IEEE/CVF conference on computer
  vision and pattern recognition. pp. 11621--11631 (2020)

\bibitem{chan2023generative}
Chan, E.R., Nagano, K., Chan, M.A., Bergman, A.W., Park, J.J., Levy, A.,
  Aittala, M., De~Mello, S., Karras, T., Wetzstein, G.: Generative novel view
  synthesis with 3d-aware diffusion models. In: Proceedings of the IEEE/CVF
  International Conference on Computer Vision. pp. 4217--4229 (2023)

\bibitem{chen2025videodepthanythingconsistent}
Chen, S., Guo, H., Zhu, S., Zhang, F., Huang, Z., Feng, J., Kang, B.: Video
  depth anything: Consistent depth estimation for super-long videos.
  arXiv:2501.12375  (2025)

\bibitem{chen2024omnire}
Chen, Z., Yang, J., Huang, J., de~Lutio, R., Esturo, J.M., Ivanovic, B.,
  Litany, O., Gojcic, Z., Fidler, S., Pavone, M., Song, L., Wang, Y.: Omnire:
  Omni urban scene reconstruction. arXiv preprint arXiv:2408.16760  (2024)

\bibitem{Dosovitskiy17CARLA}
Dosovitskiy, A., Ros, G., Codevilla, F., L{\'o}pez, A., Koltun, V.: Carla: An
  open urban driving simulator. In: Proc. CoRL (PMLR) (2017)

\bibitem{fan2025freesim}
Fan, L., Zhang, H., Wang, Q., Li, H., Zhang, Z.: Freesim: Toward free-viewpoint
  camera simulation in driving scenes. In: Proceedings of the Computer Vision
  and Pattern Recognition Conference. pp. 12004--12014 (2025)

\bibitem{fischer2024dynamic}
Fischer, T., Kulhanek, J., Bul{\`o}, S.R., Porzi, L., Pollefeys, M.,
  Kontschieder, P.: Dynamic 3d gaussian fields for urban areas. In: The
  Thirty-eighth Annual Conference on Neural Information Processing Systems
  (2024)

\bibitem{Gaidon2016VKITTI}
Gaidon, A., Wang, Q., Cabon, Y., Vig, E.: Virtual worlds as proxy for
  multi-object tracking analysis. In: Proc. CVPR (2016)

\bibitem{2025DiST}
Guo, J., Ding, Y., Chen, X., Chen, S., Li, B., Zou, Y., Lyu, X., Tan, F., Qi,
  X., Li, Z.: Dist-4d: Disentangled spatiotemporal diffusion with metric depth
  for 4d driving scene generation  (2025)

\bibitem{ho2020denoising}
Ho, J., Jain, A., Abbeel, P.: Denoising diffusion probabilistic models.
  Advances in neural information processing systems  \textbf{33},  6840--6851
  (2020)

\bibitem{huang2024s3gaussian}
Huang, N., Wei, X., Zheng, W., An, P., Lu, M., Zhan, W., Tomizuka, M., Keutzer,
  K., Zhang, S.: S3gaussian: Self-supervised street gaussians for autonomous
  driving. arXiv preprint arXiv:2405.20323  (2024)

\bibitem{kerbl20233d}
Kerbl, B., Kopanas, G., Leimk{\"u}hler, T., Drettakis, G.: 3d gaussian
  splatting for real-time radiance field rendering. ACM Transactions on
  Graphics  \textbf{42}(4) (2023)

\bibitem{Li2024XLD}
Li, H., Yuan, M., Zhang, Y., Wu, C., Zhao, C., Song, C., Feng, H., Ding, E.,
  Zhang, D., Wang, J.: Xld: A cross-lane dataset for benchmarking novel driving
  view synthesis. arXiv:2406.18360  (2024)

\bibitem{liang2025distillingmonocularfoundationmodel}
Liang, Y., Hu, Y., Shao, W., Fu, Y.: Distilling monocular foundation model for
  fine-grained depth completion (2025)

\bibitem{liu2023zero}
Liu, R., Wu, R., Van~Hoorick, B., Tokmakov, P., Zakharov, S., Vondrick, C.:
  Zero-1-to-3: Zero-shot one image to 3d object. In: Proceedings of the
  IEEE/CVF international conference on computer vision. pp. 9298--9309 (2023)

\bibitem{liu2022convnet}
Liu, Z., Mao, H., Wu, C.Y., Feichtenhofer, C., Darrell, T., Xie, S.: A convnet
  for the 2020s. In: Proceedings of the IEEE/CVF conference on computer vision
  and pattern recognition. pp. 11976--11986 (2022)

\bibitem{melas2023realfusion}
Melas-Kyriazi, L., Laina, I., Rupprecht, C., Vedaldi, A.: Realfusion: 360deg
  reconstruction of any object from a single image. In: Proceedings of the
  IEEE/CVF conference on computer vision and pattern recognition. pp.
  8446--8455 (2023)

\bibitem{mildenhall2020nerf}
Mildenhall, B., Srinivasan, P.P., Tancik, M., Barron, J.T., Ramamoorthi, R.,
  Ng, R.: Nerf: Representing scenes as neural radiance fields for view
  synthesis. In: ECCV (2020)

\bibitem{ni2025recondreamer}
Ni, C., Zhao, G., Wang, X., Zhu, Z., Qin, W., Huang, G., Liu, C., Chen, Y.,
  Wang, Y., Zhang, X., et~al.: Recondreamer: Crafting world models for driving
  scene reconstruction via online restoration. In: Proceedings of the Computer
  Vision and Pattern Recognition Conference. pp. 1559--1569 (2025)

\bibitem{rematas2022urban}
Rematas, K., Liu, A., Srinivasan, P.P., Barron, J.T., Tagliasacchi, A.,
  Funkhouser, T., Ferrari, V.: Urban radiance fields. In: CVPR (2022)

\bibitem{ren2025cosmos}
Ren, X., Lu, Y., Cao, T., Gao, R., Huang, S., Sabour, A., Shen, T., Pfaff, T.,
  Wu, J.Z., Chen, R., et~al.: Cosmos-drive-dreams: Scalable synthetic driving
  data generation with world foundation models. arXiv preprint arXiv:2506.09042
   (2025)

\bibitem{ren2025gen3c}
Ren, X., Shen, T., Huang, J., Ling, H., Lu, Y., Nimier-David, M., M{\"u}ller,
  T., Keller, A., Fidler, S., Gao, J.: Gen3c: 3d-informed world-consistent
  video generation with precise camera control. In: Proceedings of the Computer
  Vision and Pattern Recognition Conference. pp. 6121--6132 (2025)

\bibitem{Richter2016PlayingForData}
Richter, S.R., Vineet, V., Roth, S., Koltun, V.: Playing for data: Ground truth
  from computer games. In: Proc. ECCV (2016)

\bibitem{rombach2022high}
Rombach, R., Blattmann, A., Lorenz, D., Esser, P., Ommer, B.: High-resolution
  image synthesis with latent diffusion models. In: Proceedings of the IEEE/CVF
  conference on computer vision and pattern recognition. pp. 10684--10695
  (2022)

\bibitem{Ros2016SYNTHIA}
Ros, G., Sellart, L., Materzynska, J., V{\'a}zquez, D., L{\'o}pez, A.M.: The
  synthia dataset: A large collection of synthetic images for semantic
  segmentation of urban scenes. In: Proc. CVPR (2016)

\bibitem{zeronvs}
Sargent, K., Li, Z., Shah, T., Herrmann, C., Yu, H.X., Zhang, Y., Chan, E.R.,
  Lagun, D., Fei-Fei, L., Sun, D., Wu, J.: {ZeroNVS}: Zero-shot 360-degree view
  synthesis from a single real image. CVPR, 2024  (2023)

\bibitem{song2020denoising}
Song, J., Meng, C., Ermon, S.: Denoising diffusion implicit models. arXiv
  preprint arXiv:2010.02502  (2020)

\bibitem{sun2020scalability}
Sun, P., Kretzschmar, H., Dotiwalla, X., Chouard, A., Patnaik, V., Tsui, P.,
  Guo, J., Zhou, Y., Chai, Y., Caine, B., et~al.: Scalability in perception for
  autonomous driving: Waymo open dataset. In: CVPR (2020)

\bibitem{tancik2022block}
Tancik, M., Casser, V., Yan, X., Pradhan, S., Mildenhall, B., Srinivasan, P.P.,
  Barron, J.T., Kretzschmar, H.: Block-nerf: Scalable large scene neural view
  synthesis. In: Proceedings of the IEEE/CVF Conference on Computer Vision and
  Pattern Recognition. pp. 8248--8258 (2022)

\bibitem{tonderski2024neurad}
Tonderski, A., Lindstr{\"o}m, C., Hess, G., Ljungbergh, W., Svensson, L.,
  Petersson, C.: Neurad: Neural rendering for autonomous driving. In:
  Proceedings of the IEEE/CVF Conference on Computer Vision and Pattern
  Recognition. pp. 14895--14904 (2024)

\bibitem{turki2022mega}
Turki, H., Ramanan, D., Satyanarayanan, M.: Mega-nerf: Scalable construction of
  large-scale nerfs for virtual fly-throughs. In: CVPR (2022)

\bibitem{turki2023suds}
Turki, H., Zhang, J.Y., Ferroni, F., Ramanan, D.: Suds: Scalable urban dynamic
  scenes. In: Proceedings of the IEEE/CVF Conference on Computer Vision and
  Pattern Recognition. pp. 12375--12385 (2023)

\bibitem{wan2025wanopenadvancedlargescale}
Wan, T., Wang, A., Ai, B., Wen, B., Mao, C., Xie, C.W., Chen, D., Yu, F., Zhao,
  H., Yang, J., Zeng, J., Wang, J., Zhang, J., Zhou, J., Wang, J., Chen, J.,
  Zhu, K., Zhao, K., Yan, K., Huang, L., Feng, M., Zhang, N., Li, P., Wu, P.,
  Chu, R., Feng, R., Zhang, S., Sun, S., Fang, T., Wang, T., Gui, T., Weng, T.,
  Shen, T., Lin, W., Wang, W., Wang, W., Zhou, W., Wang, W., Shen, W., Yu, W.,
  Shi, X., Huang, X., Xu, X., Kou, Y., Lv, Y., Li, Y., Liu, Y., Wang, Y.,
  Zhang, Y., Huang, Y., Li, Y., Wu, Y., Liu, Y., Pan, Y., Zheng, Y., Hong, Y.,
  Shi, Y., Feng, Y., Jiang, Z., Han, Z., Wu, Z.F., Liu, Z.: Wan: Open and
  advanced large-scale video generative models. arXiv preprint arXiv:2503.20314
   (2025)

\bibitem{wang2024freevsgenerativeviewsynthesis}
Wang, Q., Fan, L., Wang, Y., Chen, Y., Zhang, Z.: Freevs: Generative view
  synthesis on free driving trajectory (2024)

\bibitem{wang2023drivedreamer}
Wang, X., Zhu, Z., Huang, G., Chen, X., Zhu, J., Lu, J.: Drivedreamer: Towards
  real-world-driven world models for autonomous driving. arXiv preprint
  arXiv:2309.09777  (2023)

\bibitem{wu2023mapnerf}
Wu, C., Sun, J., Shen, Z., Zhang, L.: Mapnerf: Incorporating map priors into
  neural radiance fields for driving view simulation. In: 2023 IEEE/RSJ
  International Conference on Intelligent Robots and Systems (IROS). pp.
  7082--7088. IEEE (2023)

\bibitem{wu20234dgaussians}
Wu, G., Yi, T., Fang, J., Xie, L., Zhang, X., Wei, W., Liu, W., Tian, Q.,
  Xinggang, W.: 4d gaussian splatting for real-time dynamic scene rendering.
  arXiv preprint arXiv:2310.08528  (2023)

\bibitem{wu2024reconfusion}
Wu, R., Mildenhall, B., Henzler, P., Park, K., Gao, R., Watson, D., Srinivasan,
  P.P., Verbin, D., Barron, J.T., Poole, B., et~al.: Reconfusion: 3d
  reconstruction with diffusion priors. In: Proceedings of the IEEE/CVF
  Conference on Computer Vision and Pattern Recognition. pp. 21551--21561
  (2024)

\bibitem{wu2023mars}
Wu, Z., Liu, T., Luo, L., Zhong, Z., Chen, J., Xiao, H., Hou, C., Lou, H.,
  Chen, Y., Yang, R., Huang, Y., Ye, X., Yan, Z., Shi, Y., Liao, Y., Zhao, H.:
  Mars: An instance-aware, modular and realistic simulator for autonomous
  driving. CICAI  (2023)

\bibitem{xie2023s}
Xie, Z., Zhang, J., Li, W., Zhang, F., Zhang, L.: S-nerf: Neural radiance
  fields for street views. arXiv preprint arXiv:2303.00749  (2023)

\bibitem{yan2024street}
Yan, Y., Lin, H., Zhou, C., Wang, W., Sun, H., Zhan, K., Lang, X., Zhou, X.,
  Peng, S.: Street gaussians: Modeling dynamic urban scenes with gaussian
  splatting. In: ECCV (2024)

\bibitem{2024StreetCrafter}
Yan, Y., Xu, Z., Lin, H., Jin, H., Guo, H., Wang, Y., Zhan, K., Lang, X., Bao,
  H., Zhou, X.: Streetcrafter: Street view synthesis with controllable video
  diffusion models  (2024)

\bibitem{yang2023emernerf}
Yang, J., Ivanovic, B., Litany, O., Weng, X., Kim, S.W., Li, B., Che, T., Xu,
  D., Fidler, S., Pavone, M., et~al.: Emernerf: Emergent spatial-temporal scene
  decomposition via self-supervision. arXiv preprint arXiv:2311.02077  (2023)

\bibitem{yang2023unisim}
Yang, Z., Chen, Y., Wang, J., Manivasagam, S., Ma, W.C., Yang, A.J., Urtasun,
  R.: Unisim: A neural closed-loop sensor simulator. In: Proceedings of the
  IEEE/CVF Conference on Computer Vision and Pattern Recognition. pp.
  1389--1399 (2023)

\bibitem{yang2023gs4d}
Yang, Z., Yang, H., Pan, Z., Zhu, X., Zhang, L.: Real-time photorealistic
  dynamic scene representation and rendering with 4d gaussian splatting. arXiv
  preprint arXiv 2310.10642  (2023)

\bibitem{yang2023deformable3dgs}
Yang, Z., Gao, X., Zhou, W., Jiao, S., Zhang, Y., Jin, X.: Deformable 3d
  gaussians for high-fidelity monocular dynamic scene reconstruction. arXiv
  preprint arXiv:2309.13101  (2023)

\bibitem{yu2024sgd}
Yu, Z., Wang, H., Yang, J., Wang, H., Xie, Z., Cai, Y., Cao, J., Ji, Z., Sun,
  M.: Sgd: Street view synthesis with gaussian splatting and diffusion prior.
  arXiv preprint arXiv:2403.20079  (2024)

\bibitem{zhao2024drivedreamer4dworldmodelseffective}
Zhao, G., Ni, C., Wang, X., Zhu, Z., Zhang, X., Wang, Y., Huang, G., Chen, X.,
  Wang, B., Zhang, Y., Mei, W., Wang, X.: Drivedreamer4d: World models are
  effective data machines for 4d driving scene representation (2024)

\bibitem{zhou2024hugs}
Zhou, H., Shao, J., Xu, L., Bai, D., Qiu, W., Liu, B., Wang, Y., Geiger, A.,
  Liao, Y.: Hugs: Holistic urban 3d scene understanding via gaussian splatting.
  In: Proceedings of the IEEE/CVF Conference on Computer Vision and Pattern
  Recognition. pp. 21336--21345 (2024)

\bibitem{zhou2024drivinggaussian}
Zhou, X., Lin, Z., Shan, X., Wang, Y., Sun, D., Yang, M.H.: Drivinggaussian:
  Composite gaussian splatting for surrounding dynamic autonomous driving
  scenes. In: Proceedings of the IEEE/CVF Conference on Computer Vision and
  Pattern Recognition. pp. 21634--21643 (2024)

\end{thebibliography}
